\documentclass[sigconf]{acmart}
\usepackage[font=small,skip=2pt]{caption} 
\usepackage{subcaption}
\usepackage{pgffor}
\usepackage[toc,page,header]{appendix}
\usepackage{minitoc}
\usepackage{wrapfig}
\usepackage[utf8]{inputenc} 
\usepackage[T1]{fontenc}    
\usepackage{hyperref}       
\usepackage{url}            
\usepackage{booktabs}       
\usepackage{amsfonts}       
\usepackage{nicefrac}       
\usepackage{microtype}      
\usepackage{xcolor}         
\usepackage{amsmath}
\usepackage{algorithmicx}
\usepackage{algorithm}
\usepackage{algpseudocode}
\usepackage{booktabs}
\usepackage{multirow}
\usepackage{graphicx}
\usepackage{float}
\usepackage{enumitem}
\usepackage{tcolorbox}
\usepackage{graphicx}   
\usepackage{colortbl}   
\AtBeginDocument{%
  }

\copyrightyear{2025} 
\acmYear{2025} 
\setcopyright{cc}
\setcctype{by-nc-sa}
\acmConference[CIKM '25]{Proceedings of the 34th ACM International Conference on Information and Knowledge Management}{November 10--14, 2025}{Seoul, Republic of Korea}
\acmBooktitle{Proceedings of the 34th ACM International Conference on Information and Knowledge Management (CIKM '25),  November 10--14, 2025, Seoul, Republic of Korea}
\acmDOI{10.1145/3746252.3760830}
\acmISBN{979-8-4007-2040-6 /2025 /11}

\begin{document}

\title{ASAP: Unsupervised Post-training with Label Distribution Shift Adaptive Learning Rate
}



\author{Heewon Park}
\email{heewon012@soongsil.ac.kr}
\orcid{0009-0006-0446-3151}
\authornote{H. Park and M. Joe contributed equally.}
\affiliation{%
  \institution{Soongsil University}
  \state{Seoul}
  \country{South Korea}
}
\author{Mugon Joe}
\authornotemark[1]
\email{mugon@soongsil.ac.kr}
\orcid{0009-0003-4111-491X}
\affiliation{%
  \institution{Soongsil University}
  \state{Seoul}
  \country{South Korea}
}
\author{Miru Kim}
\email{mirukim00@soongsil.ac.kr}
\orcid{0000-0002-5394-4780}
\affiliation{%
  \institution{Soongsil University}
  \state{Seoul}
  \country{South Korea}
}

\author{Minhae Kwon}
\authornote{Corresponding author.
All authors are with the Department of Intelligent Semiconductors, 
and M. Kwon is also affiliated with the School of Electronic Engineering.}
\email{minhae@ssu.ac.kr}
\affiliation{%
  \institution{Soongsil University}
  \state{Seoul}
  \country{South Korea}
}

\renewcommand{\shortauthors}{Heewon Park, Mugon Joe, Miru Kim and Minhae Kwon}

\begin{abstract}
In real-world applications, machine learning models face \textit{online label shift}, where label distributions change over time. Effective adaptation requires careful learning rate selection: too low slows adaptation and too high causes instability. We propose \textit{ASAP} (\textbf{A}daptive \textbf{S}hift \textbf{A}ware \textbf{P}ost-training), which dynamically adjusts the learning rate by computing the cosine distance between current and previous unlabeled outputs and mapping it within a bounded range. ASAP requires no labels, model ensembles, or past inputs, using only the previous softmax output for fast, lightweight adaptation. Experiments across multiple datasets and shift scenarios show ASAP consistently improves accuracy and efficiency, making it practical for unsupervised model adaptation.
\end{abstract}




\maketitle

\section{Introduction}

\begin{figure}
\centering
\vspace{0.2cm}
\includegraphics[width=\linewidth]{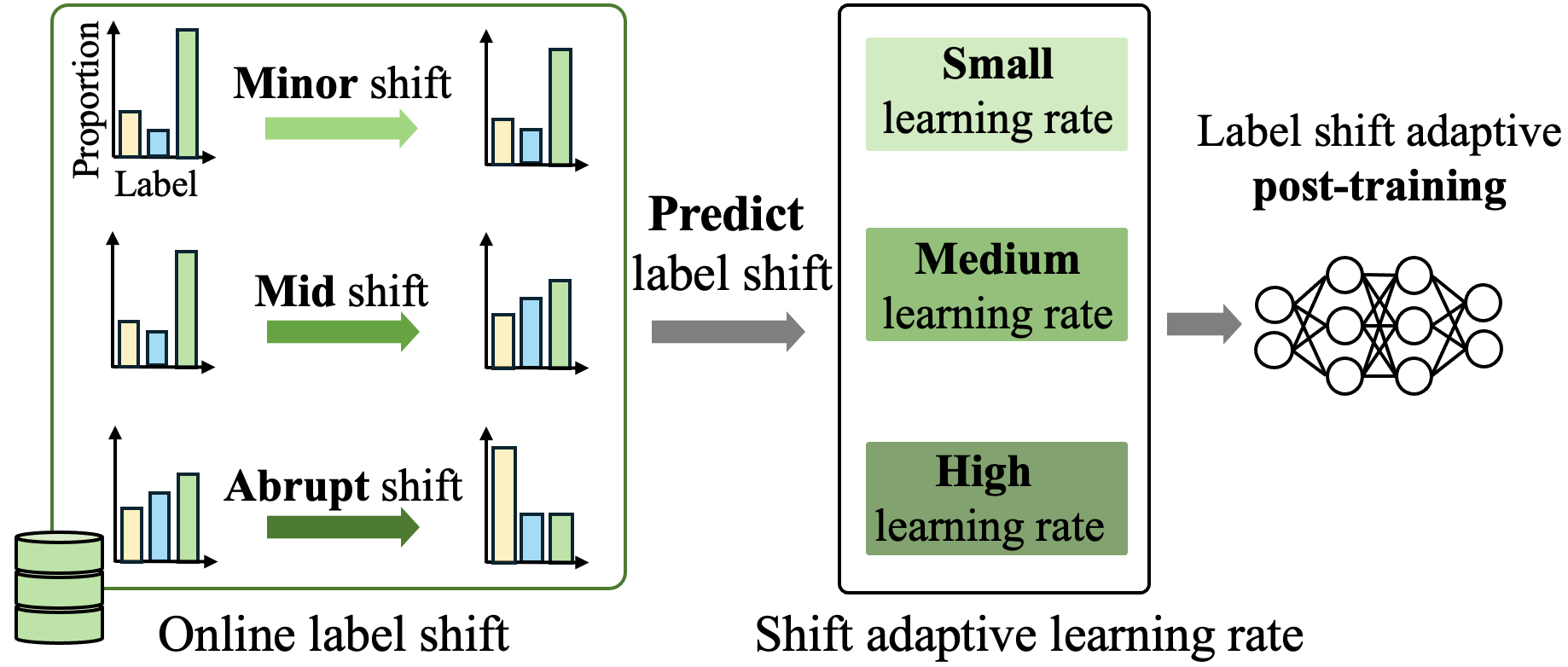} 
\hfill  
\vspace{-0.2cm}
\caption{Overview of the ASAP. ASAP estimates the degree of online label shift and assigns a learning rate accordingly: smaller shifts are assigned lower learning rates to maintain stability, while larger shifts receive higher learning rates to enhance adaptability.}
\vspace{-0.3cm}
\label{fig:overview}
\vspace{-0.3cm}
\end{figure}

Machine learning models in practical deployment settings face data streams with online label shift, where label distribution shifts occur gradually or abruptly over time~\cite{OLS1,OLS2,OLSFL2,OLSFL3,Shift-CIKM5,Shift-CIKM6,Shift-CIKM8,Shift-NeurIPS4, Shift-NeurIPS5,Shift-NeurIPS6,Shift-NeurIPS7,Shift-NeurIPS8,Shift-NeurIPS9}. Each incoming unlabeled data may exhibit changes in the underlying label distribution due to factors such as shifting user behavior, seasonal trends, or context-dependent inputs~\cite{Shift-AAAI1,Shift-AAAI2,Shift-AAAI3,Shift-AAAI4,Shift-AAAI5,Shift-CIKM3,Shift-CIKM4}. These changes can severely degrade the performance of models that were pre-trained on a static distribution, highlighting the need for continuous post-training during deployment~\cite{Shift-CIKM1,Shift-CIKM2,Shift-NeurIPS1,Shift-NeurIPS2,Shift-NeurIPS3,OLS3,OLS4,OLS5,Shift-ICML1,Shift-ICML2,Shift-ICML3,Shift-ICML4,Shift-ICML5}.

To address this issue, unsupervised post-training methods have been developed, allowing models to update without ground-truth labels~\cite{ATLAS,FTH,Shift-AAAI1,Shift-ICML1}. These methods typically estimate the distribution of current data based on model predictions and adjust parameters using the estimated distribution. However, most of them rely on fixed learning rates or complex mechanisms such as ensembles, sliding-window memory, or historical regularization. These techniques often suffer from high computational costs and are less responsive to sudden distribution changes.


One key factor that significantly influences post-training performance is the choice of learning rate. It governs how quickly the model reacts to distribution changes. For example, a small learning rate results in slow adaptation to the shift, while an large learning rate leads to unstable updates and potential forgetting of useful prior knowledge. Existing methods often rely on a fixed learning rate or select from a finite candidate set~\cite{erven2011adaptive,yang2022dltta,ATLAS}, which constrains adaptability to diverse shift patterns.

To overcome these limitations, we propose a simple yet effective method, ASAP. At each timestep, ASAP adjusts the learning rate based on how much the model’s predictions change between consecutive unlabeled data stream. It quantifies this change using the cosine distance between model's outputs and maps the result to a continuous learning rate within a bounded range. This adaptive scheduling allows for smooth and responsive adjustment to distribution changes. The learning rate remains low when predictions are stable and increases in response to the magnitude of detected distribution shifts. Despite its simplicity, ASAP consistently achieves superior performance across diverse datasets and shift scenarios, while converging faster than competitive baselines.

Our contributions are summarized as follows.
\begin{itemize}
    \item We propose a method to estimate the label distribution shift in online settings using the cosine distance between model predictions on consecutive unlabeled data samples.
    \item We propose a learning rate adjustment strategy that continuously tunes the learning rate based on the estimated magnitude of distribution shift, rather than relying on fixed or discrete candidate values.
    \item We adopt a risk estimation framework based on pseudo-label distributions to enable fully unsupervised model updates.
    \item We evaluate our method on four datasets and four types of label shift, consistently outperforming five competitive baselines in both accuracy and time efficiency.
\end{itemize}

\section{Preliminaries}
\label{preliminaries}
\vspace{-0.1cm}

\subsection{Online Label Shift}

For the pre-training phase ($t=0$), let $\theta$ denote a model trained on a pre-collected labeled data set $(\mathbf{x}^0, \mathbf{y}^0) \sim \mathbf{P}_{\mathbf{y}^0}$.  
Here, $\mathbf{P}_{\mathbf{y}^0}$ is the class distribution at time $t = 0$, where each label $\mathbf{y}^0$ is a class index $c \in \mathcal{C}$.  
The $c$-th element of $\mathbf{P}_{\mathbf{y}^0}$, denoted $[\mathbf{P}_{\mathbf{y}^0}]_c$, represents the proportion of class $c$ in the initial training labels.

At online post-training phase ($0 < t \le T$), the model encounters a sequential data stream $\{\mathbf{x}^1, \mathbf{x}^2,\dots, \mathbf{x}^T\}$, where each data $\mathbf{x}^t$ arrives without corresponding labels.  
We assume the label distribution $\mathbf{P}_{\mathbf{y}^t}$ underlying each data may shift over time, while the conditional distribution $\mathbf{P}(\mathbf{x} \mid \mathbf{y})$ remains stationary.  
This online label shift setting can be modeled using a convex interpolation as follows.
\begin{equation}
\label{eqn:label_shift_central}
\mathbf{P}_{\mathbf{y}^t} = (1 - \alpha(t))\,\mathbf{P}_{\mathbf{y}^0} + \alpha(t)\,\mathbf{P}_{\mathbf{y}^T}, \quad \alpha(t) \in [0, 1]
\end{equation}
Here, $\alpha(t)$ governs the degree of shift from the initial label distribution $\mathbf{P}_{\mathbf{y}^0}$ toward a target distribution $\mathbf{P}_{\mathbf{y}^T}$. The model $\theta$ must be adapted to reflect the evolving distribution $\mathbf{P}_{\mathbf{y}^t}$ at each time step $t$, despite lacking access to ground-truth labels.

\subsection{Learning Objective with Unsupervised Risk Estimation}
\label{section:prelim}

To support model updates with unlabeled data, we adopt an unsupervised risk estimation framework. The optimal objective with true label at time $t$ is to minimize the expected risk as follows.
\begin{equation}
\label{eq:risk_def}
\mathcal{R}^t(\theta) := \mathbb{E}_{(\mathbf{x}^t, \mathbf{y}^t) \sim \mathbf{P}_{\mathbf{y}^t}} \left[ \mathcal{L}( \mathcal{F}(\theta; \mathbf{x}^t), \mathbf{y}^t) \right] = \sum_{c \in \mathcal{C}} [\mathbf{P}_{\mathbf{y}^t}]_c \cdot \mathcal{R}^{t,c}(\theta)
\end{equation}
Here, $\mathcal{F}(\cdot)$ denotes the model’s softmax output, and $\mathcal{R}^{t,c}(\theta)$ represents the class-wise risk for class $c$, defined as the expected loss over inputs with label $c$.

Under the label shift assumption, class-conditional risks remain stable across time. Thus, the risk $\mathcal{R}^t(\theta)$ at time $t$ can be approximated using class-wise risks from the pre-training phase. 
\begin{equation}
\label{eq:risk_decomp_central}
\mathcal{R}^t(\theta) \approx \sum_{c \in \mathcal{C}} [\mathbf{P}_{\mathbf{y}^t}]_c \cdot \mathcal{R}^{0,c}(\theta)
\end{equation}
Since the true label distribution $\mathbf{P}_{\mathbf{y}^t}$ is unknown during online adaptation, we estimate it from the model’s soft predictions.  
Let $\mathbf{P}_{\widehat{\mathbf{y}}^t}$ be the pseudo-label distribution derived from model outputs on $\mathbf{x}^t$.  
We apply Black-box Shift Estimation (BBSE)~\cite{BBSE}, which estimates the true label distribution.
\begin{equation}
\mathbf{P}_{\mathbf{y}^t} \approx \mathbf{M}^{-1} \mathbf{P}_{\widehat{\mathbf{y}}^t}
\end{equation}
Here, $\mathbf{M}$ is the confusion matrix estimated from the initial labeled data $(\mathbf{x}^0, \mathbf{y}^0)$. Using this estimate, the risk in~\eqref{eq:risk_decomp_central} can be estimated without labels.
\begin{equation}
\label{eq:risk_estimated_central}
\widehat{\mathcal{R}}^t(\theta) := \sum_{c \in \mathcal{C}} \left[ \mathbf{M}^{-1} \mathbf{P}_{\widehat{\mathbf{y}}^t} \right]_c \cdot \mathcal{R}^{0,c}(\theta)
\end{equation}
Finally, the learning objective at time $t$ is to update model parameters by minimizing the estimated unsupervised risk.
\begin{equation}
\theta^{*} = \arg\min_{\theta} \widehat{\mathcal{R}}^t(\theta)
\end{equation}
This formulation provides a foundation for updating models in an unsupervised online setting, even in the presence of label shift.

\section{Shift Estimation-based Adaptive Post-training}
\label{section:method}

\begin{algorithm}[t]
\caption{ASAP for Adaptive Post-training}
\label{alg:adaptive_update}
\begin{algorithmic}[1]
\State \textbf{Input:} Pre-trained model $\theta$, learning rate bounds $[\eta_{\text{min}}, \eta_{\text{max}}]$
\State Initialize prediction buffer: $b^{0} \leftarrow \mathcal{F}(\theta; \mathbf{x}^0)$
\For{$t = 1$ to $T$}
    \State Receive new unlabeled data sample $\mathbf{x}^t$
    \State Compute current prediction: $b^t \leftarrow \mathcal{F}(\theta; \mathbf{x}^t)$
    \State Estimate shift $\mathcal{E}^t \leftarrow 1 - \frac{ \langle b^{t-1}, b^t \rangle }{ \|b^{t-1}\|_2 \cdot \|b^t\|_2 }$
    \State Compute learning rate: $\eta^t \leftarrow \eta_{\text{min}} + \mathcal{E}^t \cdot (\eta_{\text{max}} - \eta_{\text{min}})$
    \State Estimate unsupervised risk: $\widehat{\mathcal{R}}^t(\theta)$ via~\eqref{eq:risk_estimated_central}
    \State Update model parameters: $\theta \leftarrow \theta - \eta^t \nabla_\theta \widehat{\mathcal{R}}^t(\theta)$
    \State Update prediction buffer: $b^{t-1} \leftarrow b^t$
\EndFor
\end{algorithmic}
\end{algorithm}

We propose a shift-aware post-training framework ASAP for online learning in non-stationary environments. ASAP dynamically adjusts the model’s learning rate at each timestep according to the estimated degree of label distribution shift. The core insight is to use the change in the model output distribution, based on softmax predictions, to infer the distribution change and adjust the updates accordingly, without requiring labeled data.

\subsection{Shift-aware Learning Rate Adjustment}
As the model is deployed, it receives a sequence of unlabeled data samples $\{\mathbf{x}^1, \mathbf{x}^2, \dots, \mathbf{x}^T\}$ over time. To estimate distributional shifts between time steps, we track how much the model's predictions change between adjacent data samples.
At each time step $t$, we compute softmax predictions on the current input data samples, denoted as $b^t = \mathcal{F}(\theta; \mathbf{x}^t)$, and compare them with those from the previous data samples, stored in a prediction buffer $b^{t-1}$. The predicted shift $\mathcal{E}^t$ is defined as the cosine distance between these two distributions.
\begin{equation}
\label{eq:shift}
\mathcal{E}^t := 1 - \frac{ \left\langle b^{t-1}, b^t \right\rangle }{ | b^{t-1} |_2 \cdot | b^t |_2 }
\end{equation}
The value $\mathcal{E}^t$ lies in $[0, 1]$ when the predictions are probability distributions (e.g., softmax outputs), as the cosine distance between such vectors is bounded. A smaller $\mathcal{E}^t$ implies that the model's predictions have remained stable, suggesting little distributional change. Conversely, a larger value indicates a greater shift in the underlying data distribution.

After computing the shift estimate $\mathcal{E}^t$, we update the prediction buffer: $b^{t-1} \leftarrow b^t$, so that it can be used in the next iteration. This enables continual shift tracking without requiring storage of past data or labels. We then linearly map this shift value to a learning rate within a bounded range $[\eta_{\text{min}}, \eta_{\text{max}}]$.
\begin{equation}
\label{eq:lr}
    \eta^t = \eta_{\text{min}} + \mathcal{E}^t \cdot (\eta_{\text{max}} - \eta_{\text{min}})
\end{equation}
This design allows the learning rate $\eta^t$ to respond proportionally to the estimated magnitude of distribution shift. When the predicted distributions remain similar across consecutive timesteps (i.e., $\mathcal{E}^t$ is small), the update step is modest, helping to preserve learned knowledge and avoid unnecessary changes. In contrast, when the prediction change is large (i.e., $\mathcal{E}^t$ approaches 1), a higher learning rate is selected, allowing the model to rapidly adapt to potentially significant changes in the underlying data distribution.


\subsection{Model Update with Estimated Risk}

To update the model using unlabeled data, we use the unsupervised risk estimation framework described in Section~\ref{section:prelim}. At each timestep $t$, the expected risk $\widehat{\mathcal{R}}^t(\theta)$ is estimated based on pseudo-label distributions and class-wise risk statistics from pre-training.

The model is then updated using the adaptively selected learning rate as follows.
\begin{equation}
\label{eq:model_update}
    \theta \leftarrow \theta - \eta^t \nabla_\theta \widehat{\mathcal{R}}^t(\theta)
\end{equation}
This allows the model to continuously adapt in a way that is sensitive to distribution dynamics, ensuring both responsiveness and stability across time. The procedure is summarized in Algorithm~\ref{alg:adaptive_update}.

\section{Simulation}

In this section, we introduce the simulation setups, including the datasets, online label shift modeling, and baseline adaptation methods. Also, we conduct extensive experiments to validate the effectiveness of the ASAP.

\vspace{-0.1cm}
\subsection{Simulation Setup}
We evaluate on four standard benchmarks—Tiny ImageNet~\cite{Tinyimagenet}, CIFAR-10~\cite{CIFAR10}, FMNIST~\cite{FMNIST}, and MNIST~\cite{MNIST}. To simulate online label shift dynamics, we model the class prior evolution from an initial uniform distribution $\mathbf{P}_{\mathbf{y}^0}$ set uniformly across all classes to mirror pre-training conditions to a target Dirac delta distribution $\mathbf{P}_{\mathbf{y}^T}$ concentrated on a single class selected at random. This transition is governed by a time-dependent mixing coefficient $\alpha(t)$  as in~\eqref{eqn:label_shift_central}. Four label shift patterns are implemented through distinct $\alpha(t)$ formulations:

\begin{itemize}
    \item \textbf{Linear Shift} (Lin.)~\includegraphics[height=2ex]{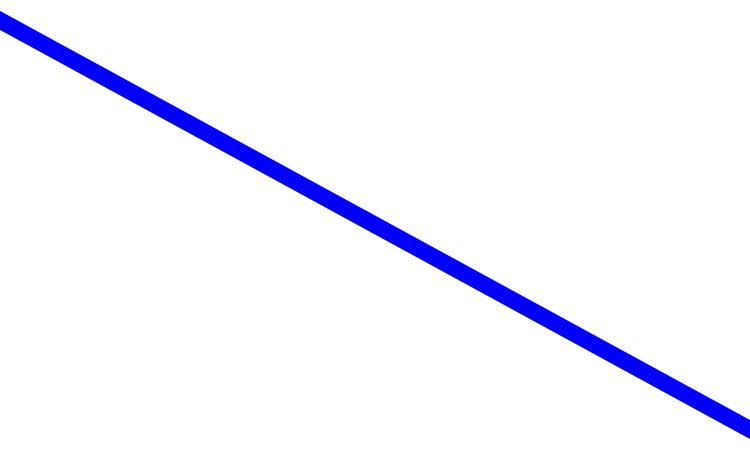}: $\alpha(t) = t/T$ produces smooth linear transitions over $T$ timesteps.
    \item \textbf{Sine Shift} (Sin.)~\includegraphics[height=2ex]{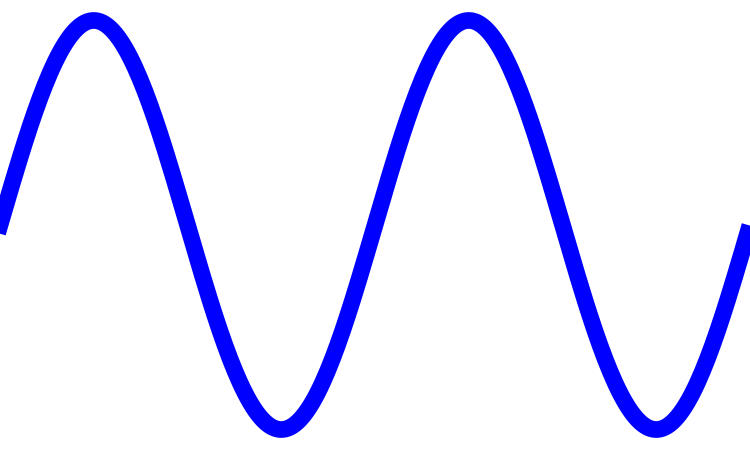}: $\alpha(t) = \sin(\pi t/\sqrt{T})$ creates periodic oscillations in class proportions.
    \item \textbf{Square Shift} (Squ.)~\includegraphics[height=2ex]{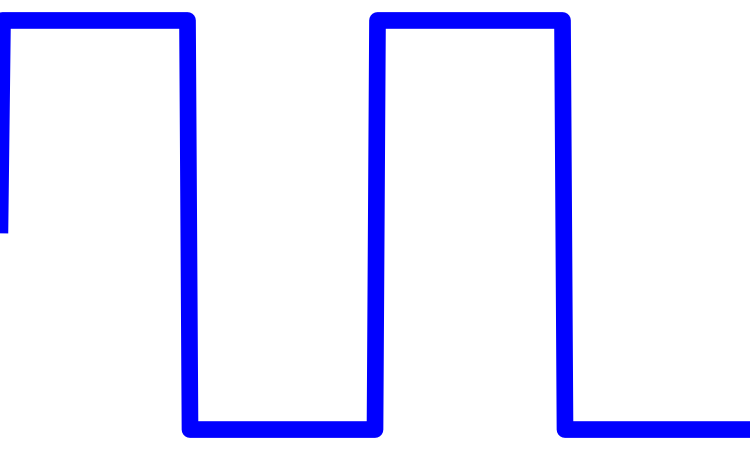}: $\alpha(t)$ alternates between 0 and 1 every $\sqrt{T}/2$ steps, generating abrupt distribution changes.
    \item \textbf{Bernoulli Shift} (Ber.)~\includegraphics[height=2ex]{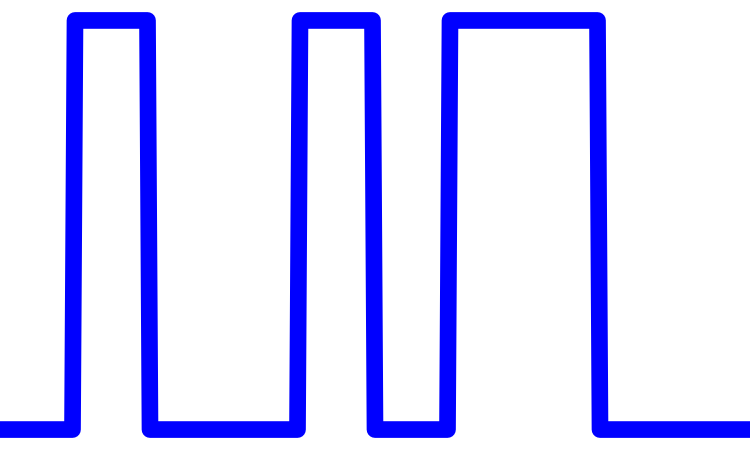}: $\alpha(t)$ flips state with probability $1/\sqrt{T}$, modeling stochastic distribution shifts.
\end{itemize}

We conduct comprehensive evaluations against five state-of-the-art algorithms for online label shift adaptation to evaluate the effectiveness of ASAP:

\begin{itemize}

\item \textbf{FTH}~\cite{FTH}: Historical averaging method that aggregates past label distributions. ASAP differs by dynamically adjusting the learning rate at each timestep, allowing for more responsive adaptation.

\item \textbf{FTFWH}~\cite{FTH}: Windowed version of FTH that focuses on recent distributions, balancing historical consistency with recent trends. ASAP instead adapts the learning rate in real time based on the estimated shift, not a fixed window size.
\item \textbf{ROGD}~\cite{FTH}: Regularized OGD with temporal constraints to prevent catastrophic forgetting. ROGD uses a fixed learning rate and does not adaptively adjust it based on distribution shift.
\item \textbf{UOGD}~\cite{ATLAS}: Unbiased OGD variant using BBSE for label distribution estimation. UOGD does not employ an adaptive learning rate mechanism and relies on a static update schedule.
\item \textbf{ATLAS}~\cite{ATLAS}: Maintains multiple base learners with different learning rates and selects the best-performing one based on recent loss. ASAP, by comparison, uses a single model and directly computes the learning rate at each time step without maintaining an ensemble.

\item \textbf{ASAP} (ours): ASAP employs cosine distance between consecutive predictions to automatically adjust learning rates, enabling stable adaptation without labels or manual tuning.
\end{itemize}



\subsection{Simulation Results}

\subsection*{Performance of Post-training}

\begin{table}[t]
\centering
\caption{Performance of post-training (average accuracy (\%))}
\label{sim:results}
\small
\setlength{\tabcolsep}{2pt}
\resizebox{\linewidth}{!}{%
\begin{tabular}{@{}cc|ccccc>{\columncolor{gray!15}}c@{}}
\toprule
\textbf{Dataset} & \textbf{Shift} 
  & FTH & FTFWH & ROGD & UOGD & ATLAS & \textbf{ASAP} (ours) \\
\midrule

\multirow{4}{*}{Tiny ImageNet} 
  & Lin. & 69.0{\tiny$\pm$13.5} & 23.6{\tiny$\pm$6.62} & 45.8{\tiny$\pm$9.8} & 45.8{\tiny$\pm$9.8} & 58.7{\tiny$\pm$10.0} & \textbf{72.4{\tiny$\pm$0.5}} \\
  & Sin. & 42.2{\tiny$\pm$10.0} & 50.7{\tiny$\pm$14.9} & 47.4{\tiny$\pm$10.0} & 47.4{\tiny$\pm$10.0} & 56.5{\tiny$\pm$10.5} & \textbf{72.3{\tiny$\pm$1.1}} \\
  & Squ. & 44.7{\tiny$\pm$10.3} & 51.9{\tiny$\pm$0.1} & 49.8{\tiny$\pm$10.6} & 49.8{\tiny$\pm$10.6} & 57.5{\tiny$\pm$11.4} & \textbf{72.2{\tiny$\pm$0.2}} \\
  & Ber. & 42.7{\tiny$\pm$12.3} & 53.6{\tiny$\pm$7.6} & 47.9{\tiny$\pm$12.8} & 47.9{\tiny$\pm$12.8} & 55.6{\tiny$\pm$13.8} & \textbf{72.5{\tiny$\pm$0.7}} \\
\midrule

\multirow{4}{*}{CIFAR-10}
  & Lin. & 66.4{\tiny$\pm$8.0} & 66.4{\tiny$\pm$8.1} & 51.1{\tiny$\pm$23.1} & 73.9{\tiny$\pm$9.9} & 60.7{\tiny$\pm$1.4} & \textbf{77.4{\tiny$\pm$3.4}} \\
  & Sin. & 51.6{\tiny$\pm$0.6} & 52.8{\tiny$\pm$1.3} & 48.7{\tiny$\pm$5.5} & 48.6{\tiny$\pm$0.8} & 60.1{\tiny$\pm$2.8} & \textbf{77.0{\tiny$\pm$3.2}} \\
  & Squ. & 42.8{\tiny$\pm$2.6} & 43.1{\tiny$\pm$2.4} & 32.7{\tiny$\pm$10.1} & 25.3{\tiny$\pm$6.2} & 61.3{\tiny$\pm$2.5} & \textbf{77.7{\tiny$\pm$3.4}} \\
  & Ber. & 19.5{\tiny$\pm$3.3} & 22.3{\tiny$\pm$4.9} & 23.5{\tiny$\pm$19.3} & 22.3{\tiny$\pm$6.7} & 59.7{\tiny$\pm$3.5} & \textbf{76.5{\tiny$\pm$2.2}} \\
\midrule

\multirow{4}{*}{FMNIST} 
  & Lin. & 71.1{\tiny$\pm$25.4} & 33.7{\tiny$\pm$10.4} & 63.1{\tiny$\pm$30.8} & 53.4{\tiny$\pm$28.3} & 86.4{\tiny$\pm$0.3} & \textbf{88.8{\tiny$\pm$0.7}} \\
  & Sin. & 71.0{\tiny$\pm$5.7} & 59.7{\tiny$\pm$19.5} & 83.0{\tiny$\pm$2.1} & 81.8{\tiny$\pm$2.5} & 86.5{\tiny$\pm$0.6} & \textbf{88.3{\tiny$\pm$0.9}} \\
  & Squ. & 75.3{\tiny$\pm$2.7} & 65.3{\tiny$\pm$11.8} & 59.8{\tiny$\pm$30.2} & 58.1{\tiny$\pm$29.7} & 86.3{\tiny$\pm$0.5} & \textbf{88.7{\tiny$\pm$0.9}} \\
  & Ber. & 74.8{\tiny$\pm$2.1} & 64.9{\tiny$\pm$13.7} & 82.2{\tiny$\pm$1.6} & 73.5{\tiny$\pm$9.1} & 86.0{\tiny$\pm$0.5} & \textbf{88.9{\tiny$\pm$1.0}} \\
\midrule

\multirow{4}{*}{MNIST} 
  & Lin. & 84.6{\tiny$\pm$24.6} & 96.5{\tiny$\pm$1.1} & 63.0{\tiny$\pm$17.7} & 81.1{\tiny$\pm$19.9} & 98.8{\tiny$\pm$0.2} & \textbf{99.2{\tiny$\pm$0.1}} \\
  & Sin. & 87.9{\tiny$\pm$7.3} & 85.5{\tiny$\pm$15.0} & 88.4{\tiny$\pm$9.3} & 77.2{\tiny$\pm$14.6} & 98.2{\tiny$\pm$0.3} & \textbf{99.2{\tiny$\pm$0.1}} \\
  & Squ. & 90.8{\tiny$\pm$6.1} & 82.5{\tiny$\pm$17.3} & 92.7{\tiny$\pm$5.1} & 83.5{\tiny$\pm$7.2} & 97.6{\tiny$\pm$0.5} & \textbf{99.2{\tiny$\pm$0.2}} \\
  & Ber. & 92.6{\tiny$\pm$5.5} & 93.4{\tiny$\pm$6.5} & 79.6{\tiny$\pm$17.5} & 76.8{\tiny$\pm$19.5} & 97.8{\tiny$\pm$0.5} & \textbf{99.2{\tiny$\pm$0.2}} \\

\bottomrule
\end{tabular}%
} \vspace{-0.5cm}
\end{table}

Table~\ref{sim:results} shows the average adaptation accuracy and standard deviation of seven online label shift adaptation methods across four datasets and four shift types, averaged over five random seeds. The wall time for each method is measured in seconds and averaged across all shift types. The ASAP consistently outperforms all baselines across all simulation results while maintaining competitive computational efficiency.

Across all datasets and shift settings, our proposed ASAP achieves the average relative improvement rates of  20.8\%, compared to the next best-performing baseline method. Furthermore, ASAP maintains stable and superior performance, with a low standard deviation of 4.27, computed across four shift types for each dataset. This is in contrast to the baselines, which suffer from highly inconsistent performance across different datasets and shift types, resulting in an average standard deviation of 13.8 across all methods. This improvement is attributed to the use of an adaptive learning rate, enabling more precise and efficient adaptation to evolving label distributions. These results demonstrate the effectiveness of the ASAP in handling online label shifts.


In computational efficiency, the wall time comparison in Table~\ref{tab:walltime} reveals that ASAP achieves superior computational efficiency across all datasets. ASAP consistently records the lowest wall time per timestep, with an average improvement of 20.3\% in adaptation speed compared to the next-best baseline. This efficiency advantage is derived from directly computing the learning rate through mathematical formulation rather than maintaining multiple candidate models. Unlike ATLAS, which requires evaluating numerous base learners with different learning rates, ASAP adapts a single model. This eliminates the computational overhead associated with ensemble maintenance and model selection, making ASAP particularly suitable for resource-constrained environments where both adaptation performance and computational efficiency are essential.

\subsection*{Parameter Sensitivity Analysis}
\begin{figure}[t]
    \centering
    \vspace{-0.1cm}
    \begin{subfigure}[b]{0.4\textwidth}
        \includegraphics[width=\linewidth]{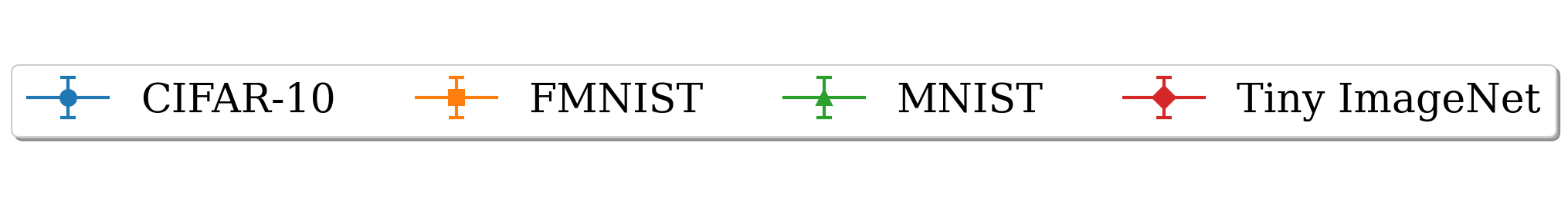}
    \end{subfigure}
    
    \vspace{-0.2cm}

    \begin{subfigure}[b]{0.22\textwidth}
        \includegraphics[width=\linewidth]{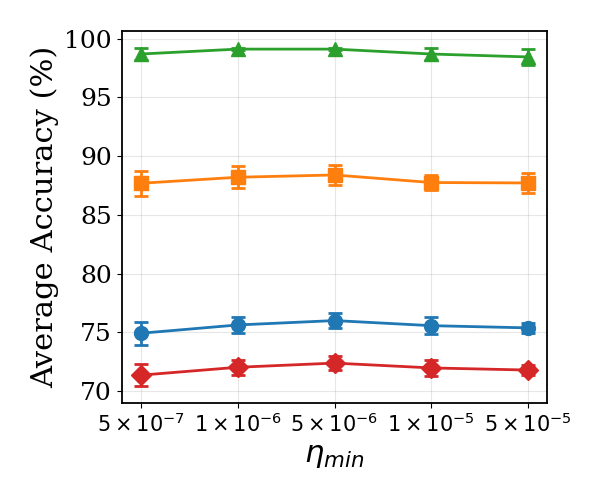}
         \vspace{-0.7cm}
        \caption{ Performance under varying \(\eta_{\text{min}}\) with fixed \(\eta_{\text{max}} = 1 \times 10^{-4}\) }
    \end{subfigure}
    \begin{subfigure}[b]{0.22\textwidth}
        \includegraphics[width=\linewidth]{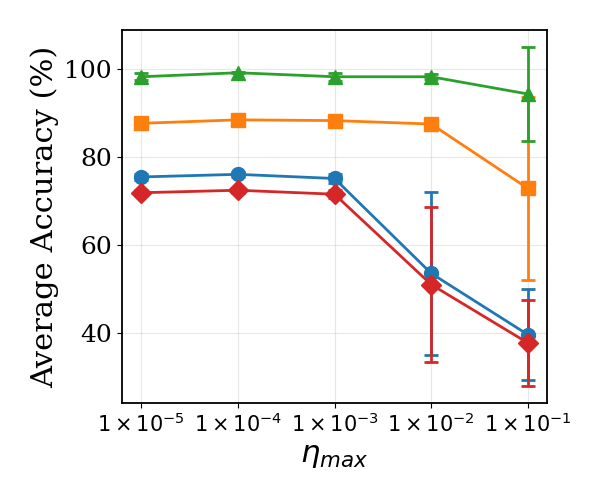}
         \vspace{-0.7cm}
        \caption{Performance under varying \(\eta_{\text{max}}\) with fixed \(\eta_{\text{min}} = 5 \times 10^{-6}\)}
    \end{subfigure}
    
    \caption{Sensitivity analysis of learning rate parameters}
    \label{fig:param}
\end{figure}

We analyze the sensitivity of ASAP to its two key hyperparameters, the minimum and maximum learning rates (\(\eta_{\text{min}}\), \(\eta_{\text{max}}\)), in Figure~\ref{fig:param}. 
For each parameter, we vary its value while keeping the other fixed: \(\eta_{\text{max}} = 1 \times 10^{-4}\) during the \(\eta_{\text{min}}\) analysis, and \(\eta_{\text{min}} = 5 \times 10^{-6}\) for the \(\eta_{\text{max}}\) analysis. 
Each plot reports average accuracy across four datasets and four label shift types, with error bars representing one standard deviation over five random seeds.

In Figure~\ref{fig:param}(a), we observe that extremely small values of \(\eta_{\text{min}} < 1 \times 10^{-6}\) lead to suboptimal adaptation, as the model fails to adjust even when the distribution changes. Conversely, large values approaching \(\eta_{\text{max}}\) (e.g., \(5 \times 10^{-5}\)) limit the modulation range of the adaptive scheduler, resulting in unstable or overly reactive updates. 
Stable and high performance is consistently observed when \(\eta_{\text{min}}\) is selected within the range \([1 \times 10^{-6}, 1 \times 10^{-5}]\).

In Figure~\ref{fig:param}(b), when analyzing \(\eta_{\text{max}}\), we find that values larger than \(1 \times 10^{-3}\) cause accuracy to deteriorate sharply. 
Setting \(\eta_{\text{max}} = 1 \times 10^{-2}\) causes an average accuracy drop of \(11.47\%\) compared to the optimal value at \(1 \times 10^{-4}\), due to overly large gradient steps destabilizing the model. 
In contrast, when \(\eta_{\text{max}}\) lies within the moderate range of \([1 \times 10^{-5}, 1 \times 10^{-3}]\), ASAP remains stable and achieves high accuracy across all datasets. 
These results confirm that ASAP is robust across a wide range of reasonable settings, as long as extreme learning rate values are avoided


\begin{table}[t]
\centering
\caption{Wall time (sec.) of each method on four datasets}
\label{tab:walltime}
\small
\setlength{\tabcolsep}{2pt}
\resizebox{\linewidth}{!}{%
\begin{tabular}{@{}c|ccccc>{\columncolor{gray!15}}c@{}}
\toprule
\textbf{Dataset} 
  & FTH & FTFWH & ROGD & UOGD & ATLAS & \textbf{ASAP} (ours) \\
\midrule

Tiny ImageNet 
  & 15.98{\tiny$\pm$0.07} & 15.43{\tiny$\pm$0.01} & 44.09{\tiny$\pm$0.11} & 44.12{\tiny$\pm$0.05} & 48.11{\tiny$\pm$0.02} & \textbf{14.54{\tiny$\pm$0.03}} \\
\midrule

CIFAR-10
  & 1.58{\tiny$\pm$0.09} & 1.62{\tiny$\pm$0.11} & 2.45{\tiny$\pm$0.12} & 2.33{\tiny$\pm$0.15} & 3.29{\tiny$\pm$0.08} & \textbf{1.12{\tiny$\pm$0.07}} \\
\midrule

FMNIST 
  & 1.52{\tiny$\pm$0.06} & 1.50{\tiny$\pm$0.07} & 2.37{\tiny$\pm$0.05} & 2.23{\tiny$\pm$0.15} & 4.20{\tiny$\pm$0.08} & \textbf{1.13{\tiny$\pm$0.09}} \\
\midrule

MNIST 
  & 1.44{\tiny$\pm$0.11} & 1.58{\tiny$\pm$0.11} & 2.29{\tiny$\pm$0.18} & 2.03{\tiny$\pm$0.14} & 7.26{\tiny$\pm$0.34} & \textbf{1.13{\tiny$\pm$0.31}} \\
\bottomrule
\end{tabular}%
}\vspace{-0.3cm}
\end{table}

\vspace{-0.1cm}
\subsection*{Learning Rate Selection Analysis}

\begin{wrapfigure}{r}{0.25\textwidth}
    \centering
    \vspace{-0.5cm}
    \includegraphics[width=\linewidth]{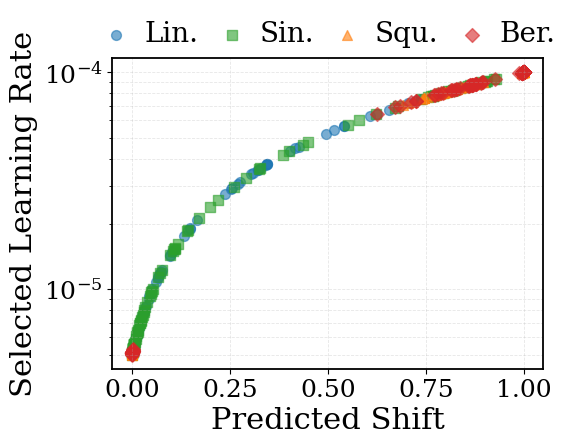}
    \caption{Selected learning rate according to predicted shift $\mathcal{E}^t$ on Tiny ImageNet}
    \label{fig:estimation}
    \vspace{-0.5cm}
\end{wrapfigure}

We visualize how the learning rate is selected at levels of estimated label shift in Figure~\ref{fig:estimation}, where the x-axis represents the shift magnitude and the y-axis denotes the chosen learning rate. The results indicate that when the estimated shift is large, the algorithm tends to select a larger learning rate, while a smaller learning rate is chosen when the estimated shift is small. 

The selected learning rate also differs according to the type of label shift. Lin. produces gradual changes in the learning rate, while Sin. results in periodic fluctuations. Squ. and Ber. lead to irregular learning rates during abrupt changes. These results demonstrate that the method adaptively adjusts the learning rate according to the underlying shift pattern. This adaptive behavior indicates that the ASAP can effectively adjust its learning rate in response to the degree of distributional change, supporting efficient adaptation to rapidly changing environments while maintaining stability when the distribution is relatively constant.

\section{Conclusion}
We proposed an unsupervised online post-training method for label distribution shift, using a shift-aware adaptive learning rate computed at each timestep without true labels. By adjusting the rate to the estimated shift magnitude, the model adapts rapidly and stably to changing distributions. Experiments across datasets and shift types show ASAP delivers superior and robust performance, maintaining high accuracy even under challenging, rapidly changing scenarios.

\section*{Acknowledgment}
This research was supported in part by National Research Foundation of Korea (NRF) grant (RS-2023-00278812, RS-2025-02214082), and in part by the Institute of Information \& communications Technology Planning \& Evaluation (IITP) grants (IITP-2025-RS-2020-II201602) funded by the Korea government (MSIT).

\newpage

\section*{GenAI Usage Disclosure}

We acknowledge the use of Generative AI (GenAI) tools in the preparation of this paper as follows:

\begin{itemize}
    \item \textbf{Writing assistance:} ChatGPT (OpenAI) was used for improving grammar, rephrasing, and refining the clarity of certain paragraphs. All substantive content and structure were authored by the authors.
    \item \textbf{Code generation:} No GenAI tools were used to generate or write code used in this study.
    \item \textbf{Data processing or analysis:} No GenAI tools were used for data processing, analysis, or result generation.
\end{itemize}

All uses of GenAI tools complied with the ACM Authorship Policy on Generative AI usage.

\bibliographystyle{unsrt}
\bibliography{reference}

\begin{thebibliography}{10}

\bibitem{OLS1}
YuYang Qian, Yong Bai, ZhenYu Zhang, Peng Zhao, and ZhiHua Zhou.
\newblock {Handling new class in online label shift}.
\newblock In {\em IEEE International Conference on Data Mining}, 2023.

\bibitem{OLS2}
Sunghyun Park, Seunghan Yang, Jaegul Choo, and Sungrack Yun.
\newblock {Label shift adapter for test-time adaptation under covariate and label shifts}.
\newblock In {\em IEEE/CVF International Conference on Computer Vision}, 2023.

\bibitem{OLSFL2}
Zhenheng Tang, Yonggang Zhang, Peijie Dong, Yiu-ming Cheung, Amelie Zhou, Bo~Han, and Xiaowen Chu.
\newblock {FuseFL: One-shot federated learning through the lens of causality with progressive model fusion}.
\newblock In {\em Advances in Neural Information Processing Systems}, 2024.

\bibitem{OLSFL3}
Naibo Wang, Yuchen Deng, Wenjie Feng, Shichen Fan, Jianwei Yin, and See-Kiong Ng.
\newblock One-shot sequential federated learning for non-iid data by enhancing local model diversity.
\newblock In {\em Proceedings of the ACM International Conference on Multimedia}, 2024.

\bibitem{Shift-CIKM5}
Hongyan Hao, Zhixuan Chu, Shiyi Zhu, Gangwei Jiang, Yan Wang, Caigao Jiang, James~Y Zhang, Wei Jiang, Siqiao Xue, and Jun Zhou.
\newblock Continual learning in predictive autoscaling.
\newblock In {\em Proceedings of the ACM International Conference on Information and Knowledge Management}, 2023.

\bibitem{Shift-CIKM6}
Kai Yao, Zixian Su, Xi~Yang, Jie Sun, and Kaizhu Huang.
\newblock Explore epistemic uncertainty in domain adaptive semantic segmentation.
\newblock In {\em Proceedings of the ACM International Conference on Information and Knowledge Management}, 2023.

\bibitem{Shift-CIKM8}
Jie Liao, Jintang Li, Liang Chen, Bingzhe Wu, Yatao Bian, and Zibin Zheng.
\newblock {SAILOR}: Structural augmentation based tail node representation learning.
\newblock In {\em Proceedings of the ACM International Conference on Information and Knowledge Management}, 2023.

\bibitem{Shift-NeurIPS4}
Xingyu Zhu, Beier Zhu, Yi~Tan, Shuo Wang, Yanbin Hao, and Hanwang Zhang.
\newblock Enhancing zero-shot vision models by label-free prompt distribution learning and bias correcting.
\newblock In {\em Advances in Neural Information Processing Systems}, 2024.

\bibitem{Shift-NeurIPS5}
Pratiksha Thaker, Amrith Setlur, Zhiwei~S Wu, and Virginia Smith.
\newblock On the benefits of public representations for private transfer learning under distribution shift.
\newblock In {\em Advances in Neural Information Processing Systems}, 2024.

\bibitem{Shift-NeurIPS6}
Zhen-Yu Zhang, Zhiyu Xie, Huaxiu Yao, and Masashi Sugiyama.
\newblock Test-time adaptation in non-stationary environments via adaptive representation alignment.
\newblock In {\em Advances in Neural Information Processing Systems}, 2024.

\bibitem{Shift-NeurIPS7}
Yarin Bar, Shalev Shaer, and Yaniv Romano.
\newblock Protected test-time adaptation via online entropy matching: A betting approach.
\newblock In {\em Advances in Neural Information Processing Systems}, 2024.

\bibitem{Shift-NeurIPS8}
Eungyeup Kim, Mingjie Sun, Christina Baek, Aditi Raghunathan, and J~Zico Kolter.
\newblock Test-time adaptation induces stronger accuracy and agreement-on-the-line.
\newblock In {\em Advances in Neural Information Processing Systems}, 2024.

\bibitem{Shift-NeurIPS9}
Yoonki Cho, Jaeyoon Kim, Woo~J Kim, Junsik Jung, and Sung-eui Yoon.
\newblock Generalizable person re-identification via balancing alignment and uniformity.
\newblock In {\em Advances in Neural Information Processing Systems}, 2024.

\bibitem{Shift-AAAI1}
Shreyas Havaldar, Jatin Chauhan, Karthikeyan Shanmugam, Jay Nandy, and Aravindan Raghuveer.
\newblock Fairness under covariate shift: Improving fairness-accuracy tradeoff with few unlabeled test samples.
\newblock In {\em Proceedings of the AAAI Conference on Artificial Intelligence}, 2024.

\bibitem{Shift-AAAI2}
Anh~T Nguyen, Lam Tran, Anh Tong, Tuan-Duy~H Nguyen, and Toan Tran.
\newblock {CASUA}l: Conditional support alignment for domain adaptation with label shift.
\newblock In {\em Proceedings of the AAAI Conference on Artificial Intelligence}, 2025.

\bibitem{Shift-AAAI3}
Dexter Neo, Stefan Winkler, and Tsuhan Chen.
\newblock {MaxEnt} loss: constrained maximum entropy for calibration under out-of-distribution shift.
\newblock In {\em Proceedings of the AAAI Conference on Artificial Intelligence}, 2024.

\bibitem{Shift-AAAI4}
Yujie Chen, Wenhui Wu, Le~Ou-Yang, Ran Wang, and Debby~D Wang.
\newblock {GeCC}: Generalized contrastive clustering with domain shifts modeling.
\newblock In {\em Proceedings of the AAAI Conference on Artificial Intelligence}, 2025.

\bibitem{Shift-AAAI5}
Yanan Wu, Zhixiang Chi, Yang Wang, Konstantinos~N Plataniotis, and Songhe Feng.
\newblock Test-time domain adaptation by learning domain-aware batch normalization.
\newblock In {\em Proceedings of the AAAI Conference on Artificial Intelligence}, 2024.

\bibitem{Shift-CIKM3}
Guoxin Chen, Yongqing Wang, Fangda Guo, Qinglang Guo, Jiangli Shao, Huawei Shen, and Xueqi Cheng.
\newblock Causality and independence enhancement for biased node classification.
\newblock In {\em Proceedings of the ACM International Conference on Information and Knowledge Management}, 2023.

\bibitem{Shift-CIKM4}
Anique Tahir, Lu~Cheng, and Huan Liu.
\newblock Fairness through aleatoric uncertainty.
\newblock In {\em Proceedings of the ACM International Conference on Information and Knowledge Management}, 2023.

\bibitem{Shift-CIKM1}
Dong Li, Chen Zhao, Minglai Shao, and Wenjun Wang.
\newblock Learning fair invariant representations under covariate and correlation shifts simultaneously.
\newblock In {\em Proceedings of the ACM International Conference on Information and Knowledge Management}, 2024.

\bibitem{Shift-CIKM2}
Naman Khetan, Sanyog Dewani, Gokul Swamy, and Vikalp Gajbhiye.
\newblock {XCapsUTL}: Cross-domain unsupervised transfer learning framework using a capsule neural network.
\newblock In {\em Proceedings of thed ACM International Conference on Information and Knowledge Management}, 2024.

\bibitem{Shift-NeurIPS1}
Yuli Slavutsky and Yuval Benjamini.
\newblock Class distribution shifts in zero-shot learning: Learning robust representations.
\newblock In {\em Advances in Neural Information Processing Systems}, 2024.

\bibitem{Shift-NeurIPS2}
Surbhi Goel, Abhishek Shetty, Konstantinos Stavropoulos, and Arsen Vasilyan.
\newblock Tolerant algorithms for learning with arbitrary covariate shift.
\newblock In {\em Advances in Neural Information Processing Systems}, 2024.

\bibitem{Shift-NeurIPS3}
Jiayun Wu, Jiashuo Liu, Peng Cui, and Steven~Z Wu.
\newblock Bridging multicalibration and out-of-distribution generalization beyond covariate shift.
\newblock In {\em Advances in Neural Information Processing Systems}, 2024.

\bibitem{OLS3}
Chuang Chen, Jiantao Shi, Mouquan Shen, Ningyun Lu, Hui Yu, Yukun Chen, and Cunsong Wang.
\newblock {Pseudo-label guided sparse deep belief network learning method for fault diagnosis of radar critical components}.
\newblock {\em IEEE Transactions on Instrumentation and Measurement}, 72:1--12, 2023.

\bibitem{OLS4}
Suresh Amalapuram, Bheemarjuna Tamma, and Sumohana Channappayya.
\newblock {SPIDER: A semi-supervised continual learning-based network intrusion detection system}.
\newblock In {\em IEEE Conference on Computer Communications}, 2024.

\bibitem{OLS5}
Dheeraj Baby, Saurabh Garg, TzuChing Yen, Sivaraman Balakrishnan, Zachary Lipton, and YuXiang Wang.
\newblock {Online label shift: Optimal dynamic regret meets practical algorithms}.
\newblock In {\em Advances in Neural Information Processing Systems}, 2024.

\bibitem{Shift-ICML1}
Yue He, Dongbai Li, Pengfei Tian, Han Yu, Jiashuo Liu, Hao Zou, and Peng Cui.
\newblock Domain-wise data acquisition to improve performance under distribution shift.
\newblock In {\em International Conference on Machine Learning}, 2024.

\bibitem{Shift-ICML2}
Tong Wei, Zhen Mao, Zi-Hao Zhou, Yuanyu Wan, and Min-Ling Zhang.
\newblock Learning label shift correction for test-agnostic long-tailed recognition.
\newblock In {\em International Conference on Machine Learning}, 2024.

\bibitem{Shift-ICML3}
Jae-Hong Lee and Joon-Hyuk Chang.
\newblock Stationary latent weight inference for unreliable observations from online test-time adaptation.
\newblock In {\em International Conference on Machine Learning}, 2024.

\bibitem{Shift-ICML4}
Dapeng Hu, Jian Liang, Xinchao Wang, and Chuan-Sheng Foo.
\newblock Pseudo-calibration: improving predictive uncertainty estimation in unsupervised domain adaptation.
\newblock In {\em International Conference on Machine Learning}, 2024.

\bibitem{Shift-ICML5}
Wei Wang, Chao Huang, Jie Wen, and Cong Wang.
\newblock Batch singular value polarization and weighted semantic augmentation for universal domain adaptation.
\newblock In {\em International Conference on Machine Learning}, 2024.

\bibitem{ATLAS}
Yong Bai, YuJie Zhang, Peng Zhao, Masashi Sugiyama, and ZhiHua Zhou.
\newblock {Adapting to online label shift with provable guarantees}.
\newblock In {\em Advances in Neural Information Processing Systems}, 2022.

\bibitem{FTH}
Ruihan Wu, Chuan Guo, Yi~Su, and Kilian Weinberger.
\newblock {Online adaptation to label distribution shift}.
\newblock In {\em Advances in Neural Information Processing Systems}, 2021.

\bibitem{erven2011adaptive}
Tim Erven, Wouter~M Koolen, Steven Rooij, and Peter Gr{\"u}nwald.
\newblock Adaptive hedge.
\newblock {\em Advances in Neural Information Processing Systems}, 2011.

\bibitem{yang2022dltta}
Hongzheng Yang, Cheng Chen, Meirui Jiang, Quande Liu, Jianfeng Cao, Pheng~Ann Heng, and Qi~Dou.
\newblock {DLTTA: Dynamic learning rate for test-time adaptation on cross-domain medical images}.
\newblock {\em IEEE Transactions on Medical Imaging}, 41(12):3575--3586, 2022.

\bibitem{BBSE}
Zachary Lipton, Yu-Xiang Wang, and Alexander Smola.
\newblock Detecting and correcting for label shift with black box predictors.
\newblock In {\em International conference on machine learning}, 2018.

\bibitem{Tinyimagenet}
Yann Le and Xuan Yang.
\newblock Tiny imagenet visual recognition challenge.
\newblock {\em CS 231N}, 7(7):3, 2015.

\bibitem{CIFAR10}
Alex Krizhevsky and Geoffrey Hinton.
\newblock Learning multiple layers of features from tiny images.
\newblock {\em Technical report, Univ. of Toronto}, 2009.

\bibitem{FMNIST}
Han Xiao, Kashif Rasul, and Roland Vollgraf.
\newblock {Fashion-MNIST: A novel image dataset for benchmarking machine learning algorithms}.
\newblock {\em arXiv preprint arXiv:1708.07747}, 2017.

\bibitem{MNIST}
Yann LeCun, Bernhard Boser, John~S Denker, Donnie Henderson, Richard~E Howard, Wayne Hubbard, and Lawrence~D Jackel.
\newblock Backpropagation applied to handwritten zip code recognition.
\newblock {\em Neural Computation}, 1(4):541--551, 1989.

\end{thebibliography}

\end{document}